\documentclass[10pt,twocolumn,letterpaper]{article}

\usepackage{wacv}
\usepackage{times}
\usepackage{epsfig}
\usepackage{graphicx}
\usepackage{amsmath}
\usepackage{amssymb}
\usepackage{array}

\usepackage{caption}
\usepackage{color}
\usepackage{subfigure}
\usepackage{epstopdf}
\usepackage{array}
\newcolumntype{P}[1]{>{\centering\arraybackslash}p{#1}}

\newcommand{\old}[1]{}
\wacvfinalcopy % *** Uncomment this line for the final submission

\def\wacvPaperID{312} % *** Enter the wacv Paper ID here

\ifwacvfinal\fi
\begin{document}

%%%%%%%%% TITLE
\title{High-speed Video from Asynchronous Camera Array}

% Authors at the same institution
%\author{First Author \hspace{2cm} Second Author \\
% Authors at different institutions
\author{Si Lu \\
Portland State University\\
{\tt\small lusi@pdx.edu}
}

\maketitle
\ifwacvfinal\thispagestyle{empty}\fi

%%%%%%%%% ABSTRACT
\begin{abstract}
   This paper presents a method for capturing high-speed video using an asynchronous camera array. Our method sequentially fires each sensor in a camera array with a small time offset and assembles captured frames into a high-speed video according to the time stamps. The resulting video, however, suffers from parallax jittering caused by the viewpoint difference among sensors in the camera array. To address this problem, we develop a dedicated novel view synthesis algorithm that transforms the video frames as if they were captured by a single reference sensor. Specifically, for any frame from a non-reference sensor, we find the two temporally neighboring frames captured by the reference sensor. Using these three frames, we render a new frame with the same time stamp as the non-reference frame but from the viewpoint of the reference sensor. Specifically, we segment these frames into super-pixels and then apply local content-preserving warping to warp them to form the new frame. We employ a multi-label Markov Random Field method to blend these warped frames. Our experiments show that our method can produce high-quality and high-speed video of a wide variety of scenes with large parallax, scene dynamics, and camera motion and outperforms several baseline and state-of-the-art approaches. 
\end{abstract}

%%%%%%%%% BODY TEXT
\vspace{-0.05in}
\section{Introduction}
\vspace{-0.05in}
\label{sec:intro}

\begin{figure*}[htb]
    \vspace{-0.30in}
    \footnotesize{
    \begin{center}
        \begin{tabular}{c}
            \hspace{-0.1in}\includegraphics[width=1.0\textwidth]{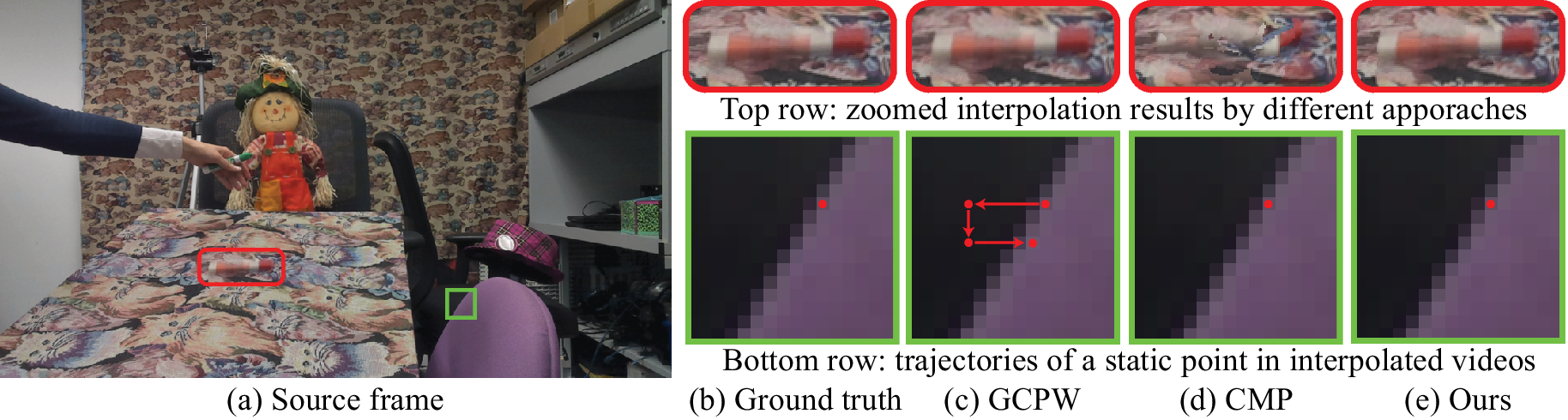}    
        \end{tabular}
        \vspace{-0.2in}
        \caption{Frame synthesis for high-speed video. (a): a source frame. (b): ground truth of the interpolated content (top) and the trajectory (bottom) of a static pixel. Global content-preserving warp (GCPW)~\cite{Liu:2009} suffers from parallax jittering in local regions as shown in (c) bottom. A state-of-the-art optical flow-based method (CMP)~\cite{hu2016efficient} cannot handle blurry object as shown in (d) top. Our method produces visually plausible results as shown in (e).\label{fig:visual_01}}
    \end{center}
    }
    \vspace{-0.325in}
\end{figure*}

%camera arrays methods not only provide a way to integrate multiple regular-speed cameras to produce high-speed video, but also enable the integration of multiple high-speed video cameras to capture video with even higher speed.

Camera arrays have been studied for decades and a variety of camera arrays have been developed in both academic labs and companies. Consumer-level camera arrays are now available. These camera arrays innovate the way of photography and videography, making many tasks easy, such as high-dynamic imaging and refocusing after the fact. 

This paper explores camera arrays for high-speed videography by sequentially firing each sensor in a camera array with a small time offset. In this way, a high-speed video can be captured by assembling the recorded frames according to their captured time. A camera array with $n$ lenses, each capturing an $m$-fps video, can record an $mn$-fps video. Compared to single-lens high-speed cameras~\cite{bloch2014effects, gulan2011experimental, ishii20102000, perry2015sanstreak}, this asynchronous camera array offers a number of advantages. First, a camera array can be made of a number of cheap normal-speed imaging sensors. Second, while the camera array method provides an economic solution for high-speed video capturing, it can be flexibly exploited to integrate multiple high-end high-frame rate cameras to capture videos with even higher frame rates. Third, a camera array can better meet the demand for high data throughput from high-speed imaging than a single-sensor camera as the processing of individual imaging sensors, such as compression, can be highly parallel. Finally, using a single-sensor camera to capture high-speed video limits the exposure time, leading to noisy images. A camera array can increase the exposure time by overlapping the explore duration between consecutive sensors.

As the imaging sensors in a camera array have small spatial baselines, the images from individual sensors must be transformed as if they were imaged from a single reference lens. Early attempts addressed this problem by treating the scene as a plane or assuming the scene is far away from the camera~\cite{etheredge2016goslow, Wilburn2004, wilburn2005high}. In this way, images from individual lenses can be transformed and aligned using a global 2D projective transformation (i.e. homography). This method cannot work well in many practical scenarios where the scene exhibits large depth variations. The resulting high-speed video typically suffers from parallax jittering. Alternatively, spatially-varying warping algorithms can be employed to warp these frames. These warping algorithms are more flexible than homography and are able to distribute distortions to visually less salient regions than the others while following a sparse set of motion displacements. These warping algorithms have been shown robust against moderate parallax in a range of applications such as image stitching~\cite{Lin2011,Zhang:14} and video stabilization~\cite{Liu:2009}. However, as these algorithms warp an image as a whole, they will produce undesirable distortions in local regions when parallax is significant, as shown in Figure~\ref{fig:visual_01} (c).  

This paper presents a novel view synthesis method that employs local spatially-varying warping and multi-label graph cuts to transform source frames as if they were captured from a common reference lens. Specifically, given any source frame and its two temporally neighboring frames captured by the reference lens, our method partitions them into super pixels. Our method then estimates dense optical flow among any of these three frames to establish correspondence between the super pixels. As dense optical flow estimation is prone to errors, not all the super pixels can be matched across these frames. To address these problems, our method merges the unmatched super pixels with the neighboring matched super pixels. Based on super pixel correspondences, our method employs a local spatially-varying warping algorithm to warp all the super pixels in the three frames to the reference locations according to its time stamp as if they were viewed by the reference camera at that moment. Linearly blending these warped super pixels from three input frames will produce ghosting artifacts. Instead, our method formulates super pixel blending as a multi-label Markov Random Field problem that properly chooses the right blending schemes for pixels to achieve visually pleasing blending results while avoiding ghosting artifacts. 

This paper contributes a method that explores the increasingly available camera array to produce high-speed video. The key enabling algorithm is a high-quality novel view synthesis algorithm that transforms video frames captured by spatially-distributed lenses as if they were captured by a common lens to avoid parallax jittering. This novel view synthesis algorithm integrates local spatially-varying warping and multi-label MRF optimization to produce a plausible novel view from multiple frames while avoiding ghosting artifacts and handling parallax. Our experiments also show that our method can produce high-quality and high-speed video of a wide variety of scenes with scene dynamics, parallax, and camera motion.

\vspace{-0.05in}
\section{Related Work}
\vspace{-0.05in}
\label{sec:related}
%%%% frame interpolation
%%%% video stabilization; here we can include stanford work
%%%% discuss the difference between our novel view synthesis and existing ones

\begin{figure*}[tb]
    \vspace{-0.05in}
    \footnotesize{
    \centering
    \begin{minipage}[b]{0.164\textwidth}
        \includegraphics[width=\textwidth]{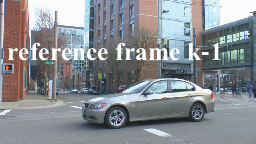} \\
        \includegraphics[width=\textwidth]{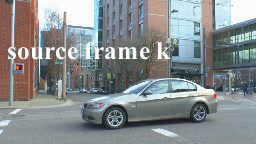}     \\ 
        \includegraphics[width=\textwidth]{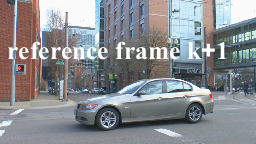} \\ [-0.75ex]
        \centering (a) Input Frames
    \end{minipage}
    \qquad \hspace{-0.33in}
    \begin{minipage}[b]{0.164\textwidth}
        \includegraphics[width=\textwidth]{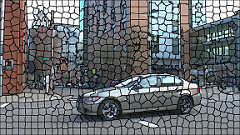} \\
        \includegraphics[width=\textwidth]{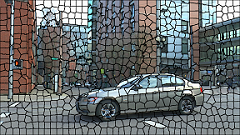}     \\
        \includegraphics[width=\textwidth]{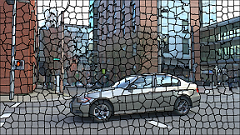} \\ [-0.75ex]
        \centering (b) SP segmentation
    \end{minipage}
    \qquad \hspace{-0.33in}
    \begin{minipage}[b]{0.164\textwidth}
        \includegraphics[width=\textwidth]{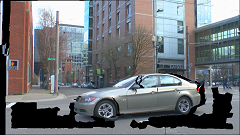} \\
        \includegraphics[width=\textwidth]{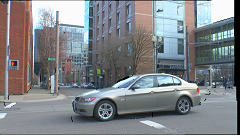}     \\
        \includegraphics[width=\textwidth]{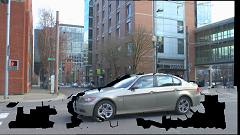} \\ [-0.75ex]
        \centering (c) Warped frames
    \end{minipage}
    \qquad \hspace{-0.33in}
    \begin{minipage}[b]{0.500\textwidth}
        \includegraphics[width=\textwidth]{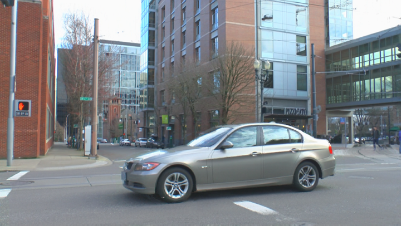} \\ [-0.75ex]
        \centering (d) Final interpolating result
    \end{minipage}
    \vspace{-0.15in}
    \caption{An example of our method. The three input frames (a), including two reference frames and one source frame, are over-segmented into superpixels (SP) (b), locally warped to the target position (c), and blended using our multi-label based optimization scheme (d).}
    \label{fig:framework}
    }
	\vspace{-0.25in}
\end{figure*}

% 1) single cameras, i: cost is high, ii: exposure time is limited, iii: data throughput
% 2) camera arrays, homography transform are used, only for distant scenes or scenes in a roughly plane
% 3) novel view synthesis, classic ones + deep learning ones
% 4) frame interpolation, classic ones + deep learning ones

This work falls into the area of frame interpolation~\cite{flow_interp_09} and novel view synthesis~\cite{Kang:2006, kopf2013image}. A complete overview of this area is out of scope of this work. We discuss the work that directly related to this paper.

\old{Our work builds upon existing methods for video frame interpolation. A typically video frame interpolation method estimates dense correspondence through optical flow between two consecutive frames and follows the optical flow to interpolate one or multiple frames in between them~\cite{flow_interp_09}. Similar to frame interpolation, our method also relies on the optical flow between two reference frames to guide frame synthesis. Our method differs from video frame interpolation in that we have extra frames that are captured at the same time as the frames to be interpolated but from different viewpoints. Therefore, we interpolate from three frames and we employ a multi-label graph cut algorithm to decide an optimal blending scheme to make optimal use of extra frames. Recently, the research on optical flow has been progressing fast~\cite{baker2004lucas,black1996robust, brox2004high, brox2011large, bruhn2005towards,chen2016full, liu2011sift, papenberg2006highly, revaud2015epicflow, sun2010secrets,vogel2013evaluation,weinzaepfel2013deepflow,werlberger2009anisotropic}, which will benefit the frame interpolation methods as well as ours. Besides the optical flow based methods, Meyer~\etal developed a phase-based interpolation method that requires no flow computation and modifies the phase difference to produce intermediate frames~\cite{meyer2015phase}. This phase approach produces impressive results; however, it is unclear how to employ this phase approach to incorporate the extra frame in our problem for better interpolation. }

A typically video frame interpolation method estimates dense correspondence using optical flow between two consecutive frames and follows the optical flow to interpolate one or multiple frames in between them~\cite{flow_interp_09}. This method, however, can fail due to the difficulty of optical-flow estimation. While traditional optical flow methods~\cite{baker2004lucas, black1996robust, brox2004high, bruhn2005towards, werlberger2009anisotropic} do not work well at object boundaries or in textureless regions, several edge-aware approaches~\cite{liu2011sift, revaud2015epicflow, weinzaepfel2013deepflow} based on edge and feature mapping have been proposed. While these methods achieve better interpolation results at object boundaries, they can not handle large motion. Later, optimization-based approaches~\cite{brox2011large, chen2016full} are developed according to different rules to deal with large motion and can generate appealing optical flow results. However, flow errors can still occur and lead to noticeable visual artifacts when using flow-based frame interpolation due to occlusion. Meyer~\etal developed phase-based interpolation methods that requires no flow computation and modifies the phase difference to produce intermediate frames~\cite{meyer2015phase,Meyer_2018_CVPR}. These phase approaches produce impressive results; however, it is unclear how to employ these approaches to incorporate the extra frame in our problem for better interpolation. Niklaus~\etal learned adaptive CNN~\cite{niklaus2017video} and content-aware CNN~\cite{Niklaus_2018_CVPR} to predict intermediate frames and achieves state-of-the-art performance. However, those method can not handle scenes with fast moving objects or large motion. Jiang~\etal proposed Super slomo~\cite{jiang2017super}, a frame work that uses a U-Net architecture to pre-compute bi-directional optical flows and fuse them to generate intermediate bi-directional optical flows at the target time stamps. Then for any time stamp, they used another U-Net to properly fuse two warped frames from both forward and backward input frames. Our method differs from those frame interpolation methods in that we have extra frames that are captured at the same time but from different viewpoints as the frames to be interpolated. Therefore, we interpolate from these frames and we employ a multi-label graph cut algorithm to decide an optimal blending scheme to make optimal use of extra frames.

Our problem can also be formulated as a video stabilization problem if we consider frames as captured by a regular camera moving along a zigzag path periodically. While a variety of video stabilization methods are now available~\cite{liu2016meshflow}, directly applying them to our problem is insufficient due to the highly patterned zigzag path, especially with large depth variation in the scene. As traditional homography stabilization approaches~\cite{matsushita2006full, Wilburn2004} fail on those video scenes, Liu~\etal ~\cite{Liu:2009} propose spatially varying warp to handle moderate depth variation. We tried to apply the content-preserving warping based approach to stabilize such a sequence. As reported in our experiment, the result still looks jittering in some local regions.

Novel view synthesis methods for camera arrays are also related to our work~\cite{liu2016meshflow, zitnick2004high}. They estimate 3D scenes from captured images, warp and blend them to create novel views. Our method is most related to Chaurasia~\etal~\cite{sigraph13warp}. This approach over-segments the input images into super-pixels and synthesizes depth for challenging regions with poor depth estimation using similar neighbouring super-pixels and warp each super-pixel individually. In this paper, we propose a frame interpolation approach to transform frames captured by an asynchronous camera array into a high-speed video. However, our input frames from different viewpoints are not taken at the same time, making it difficult to estimate depth for moving objects. Instead, we use optical flow as warping guidance and propose a validation process to eliminate bad warping guidance. We also propose a super-pixel merging scheme to propagate high quality warping guidance to nearby regions. More importantly, instead of blending all the warped frames as weighted average, our method formulates the subset selection of warped pixels for blending as a multi-labeling problem and employs a Markov Random Field method to optimize the selection to produce a visually plausible novel view.

\vspace{-0.05in}
\section{Methodology Over View}
\label{sec:method_overview}
\vspace{-0.05in}
Our method has two main steps: \textsl{optical flow guided local warp} and \textsl{graph cuts-based multi-label rendering}, as shown in Figure~\ref{fig:framework}. Given a set of alternatively captured video frames by $n$ lenses in a camera array, we consider the camera with the latest firing order at each shooting iteration as reference and other cameras as sources. We aim to transform the sources as if they were captured by the reference lens. We transform source frames one by one independently, therefore, the $n$-lens camera array problem can be simplified as a sequence of two-lens camera array ones. Without loss of generality, this section focuses on a two-lens camera array.

After we assemble the frames captured by an asynchronous two-lens camera array, we obtain a frame sequence $V= I^r_1, I^s_2, I^r_3, I^s_4, \cdots, I^r_{k-1}, I^s_k, I^r_{k+1}, \cdots$. Given two consecutive frames captured by the reference lens, $I^r_{k-1}$ and $I^r_{k+1}$ and a source frame $I^s_{k}$ between these reference frames, our goal is to generate a synthesized frame as if it was captured by the reference lens at time stamp $k$. We first compute a set of dense pixel correspondences using SparseFlow~\cite{ijcv2011opf}, including the forward and backward optical flow between the source frames ($F^{s}_{k,k-2}, F^{s}_{k,k+2}$), between the reference frames ($F^{r}_{k+1,k-1}, F^{r}_{k+1,k+3}$), from the source frame to its two temporal neighbouring reference frames ($F^{sr}_{k,k+1}, F^{sr}_{k,k-1}$), and the ones from the two reference frames back to the source frame ($F^{rs}_{k-1,k}, F^{rs}_{k+1,k}$). We then over-segment~\cite{slicSP} the three input frames into super-pixels according to both pixel intensities and the estimated flow magnitudes. Note that our approach is independent of the choice of optical flow and segmentation approaches. In addition, since we use image-based rendering guided by estimated optical flows, additional geometry information between source and reference cameras is not needed.

%\vspace{0.10in}
%\subsubsection{Optical Flow Guided Local Warp}

\noindent \textbf{Optical Flow Guided Local Warp.} Given the three input frames with estimated optical flow, we aim to warp them to a target temporal position for final rendering. Given a pixel with its estimated optical flows as well as the time stamp information, we compute its corresponding positions in other views. Thus, each pixel with good optical flow could be considered as a feature point across multiple views and could be used to guide the frame warping. However, as optical flow often contains errors, especially in occluded/dis-occluded and blurred regions, we validate each pixel's flow using a simple but effective intensity matching approach to generate an optical flow weight map $W$ for each input frame. Only pixels with high quality optical flow (large weight in $W$) are selected to guide the warp. For super-pixels with few good pixel correspondences (optical flow), a merging process is applied to merge such super-pixels to their neighbours with good pixel correspondences. This allows neighbouring super-pixels to guide the warp of such \textsl{bad} super-pixels. 

A global content-preserving warp~\cite{Liu:2009} can then be used to warp each input frame. However, this method can produce undesirable distortions when parallax is significant. We then follow the approach from Chaurasia~\etal~\cite{sigraph13warp} to warp each superpixel individually to allow plausible warping results. 

%\vspace{0.10in}
%\subsubsection{Rendering}

\noindent \textbf{Rendering.} A simple averaging approach could be used to generate the final rendering result. However, this might introduce undesirable visual artifacts, such as blurring and ghosting artifacts. While the three input frames are warped to the same temporal and spatial position, they still contain errors, such as intensity discontinuities caused by blurry objects and warping of occluded/dis-occluded regions. In addition, holes can exist due to the superpixel wise warp procedure. Thus, for each rendering pixel, the selection of its three warped sources should be carefully considered. Noticing that there are $2^3=8$ combinations of selection for each rendering pixel, we consider the selections of all rendering pixels as a labeling problem. We consider the whole rendering frame as an un-directed graph and use a graph cuts based multi-label energy minimization technique~\cite{graphcuts} with properly designed data term and smoothness term to solve the labeling problem.

Given the optimized labels for each rendering pixel, we weighted average the corresponding selected warped pixels according to optical flow validated weights $W$. Finally, we use Poisson Blending to fill the rest holes in the rendering result. In the next subsections, we will describe our \textsl{optical flow guided local warp} and \textsl{graph cuts-based multi-label rendering} in details.

\vspace{-0.05in}
\section{Optical Flow Guided Local Warp}
\label{sec:method_localwarp}
\vspace{-0.05in}
Our input is a set of three video frames, including two neighbouring reference frames ($I^r_{k-1}$, $I^r_{k+1}$) captured by the reference camera at time $t_r=0,1$, respectively, and one source frame ($I^s_k$) captured by the source camera at time $t \in (0,1)$. Our goal is to warp all the three input frames to the same temporal position as if they were imaged from the reference camera at a time spot in-between $I^r_{k-1}$ and $I^r_{k+1}$ with a temporal interpolating parameter $t \in (0,1) $. 

\subsection{Optical Flow Validation}
\label{sec:opfvalid}

\begin{figure}[t]
\vspace{-0.02in}
\footnotesize{
    \begin{center}
      \begin{tabular}{cc}
		\hspace{-0.10in}\includegraphics[width=.242\textwidth]{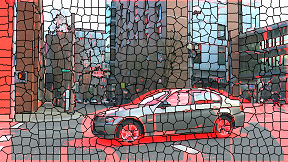}       &
        \hspace{-0.15in}\includegraphics[width=.242\textwidth]{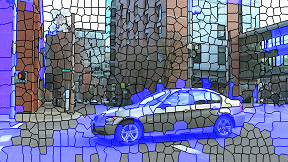}          \\ [-0.5ex]		
		\hspace{-0.13in} (a) $I^s_{k}$ overlaid with $W^{s_k}$ & \hspace{-0.15in} (b) Merged super-pixels                                                  
      \end{tabular}
	  \vspace{-0.15in}
      \caption{(a): Super-pixels near frame boundary or in occluded/dis-occluded regions often have optical flow with poor qualities (marked in red). (b): Super-pixels with low quality optical flow are merged to nearby super-pixels to form merged super-pixels (marked in blue) with enough pixels having good optical flow guidance. \label{fig:flowW_spMerge}}
    \end{center}
}
\vspace{-0.30in}
\end{figure}

Optical flow estimation results often contain errors due to the existence of blurred moving objects, occluded/dis-occluded regions and parallax effects. As shown in Figure~\ref{fig:flowW_spMerge} (a), these optical flow estimation, with poor accuracy, need to be excluded for latter warp guidance. We thus propose an intensity patch matching approach to effectively validate optical flow estimation for each pixel in the input frames. Generally, given a pixel $p^{m_1}_{i_1,j_1}$ in one of the three input frames, we first search for its corresponding pixels $p^{m_2}_{i_2,j_2},p^{m_3}_{i_3,j_3}$ in the other two frames according to the estimated optical flow. To validate the optical flow estimation for pixel $p^{m_1}_{i_1,j_1}$, we compare the corresponding patches centered at these three pixels as follows,

\vspace{-0.10in}
\small
\begin{equation}
        d_c = max(||P^{m_1}_{i_1,j_1}-P^{m_2}_{i_2,j_2}||^2 , ||P^{m_1}_{i_1,j_1}-P^{m_3}_{i_3,j_3}||^2)
\end{equation}
\normalsize

\noindent where $P^{m_n}_{i_n,j_n}$ is the patch centered at $p^{m_n}_{i_n,j_n}$ with $n=1,2,3$ and $m_n \in \{r_{k-1},r_{k+1},s_k\}$ indicating the three input frames, respectively. The optical flow validation weight is then computed as $W^m_{i,j}=e^{-(d^2_c/2\sigma^2_m)}$, where $\sigma_m$ is a pre-selected parameter. This allows us to assign high weight to a pixel if and only if it is similar to both corresponding pixels in the other two frames. To further exclude outliers, we follow the approach from Baker~\etal~\cite{flow_interp_09} to add a forward/backward optical flow check.

%\vspace{-0.075in}
\subsection{Superpixel Merging}
%\vspace{-0.02in}

Our assumption is that neighbouring super-pixels are more likely to share similar motions. We thus merge bad super-pixels to their neighbours with good flow estimation and essentially let neighbouring super-pixels to guide the warp. Here, we define \textsl{good} superpixel as a superpixel that has more than $h$ (with a default value 100) pixels with optical flow validated weights larger than a threshold $th_{sp}$ (with a default value 0.96). Specifically, for each bad superpixel $S_{b}$, we start from a queue $Q$ containing $S_{b}$ and search via expanding. At each step, we look at neighbouring super-pixels of all super-pixels in $Q$ and en-queue either a good neighbouring superpixel with good flow estimations or a bad neighbouring superpixel with the smallest motion difference to $S_{b}$ if no good neighbouring super-pixels exists. We repeat this step until at least one good superpixel is added to $Q$. All searched super-pixels in $Q$ are then combined to form a merged superpixel.

Note that our superpixel merging is only applied to source frames. This means that we try to warp all pixels in the source frames as scenes in these frames are captured at the interpolating time. For the reference frames, we simply do not warp bad pixels as they might conflict with their correspondences in the warped source frames.

\vspace{-0.075in}
\subsection{Local Content-Preserving Warp}
\vspace{-0.02in}

We now have the pre-validated optical flow and modified superpixel segmentation for all three input frames. We aim to warp each superpixel $S$ in the input frame to a target position $\hat{S}$ in the warped frame. Specifically, for each superpixel in the input frame, we construct an axis-aligned bounding box and divide it into regular mesh grids with its vertices represented as $V$. The warping of a superpixel can then be formulated as a mesh warping problem, where the unknowns are the corresponding grid vertices $\hat{V}$ in the warped frame. This mesh warping problem can then be solved as an optimization problem with a data term $E_{p,m}$ that encourages pixels to be re-projected to its potential locations for each feature point $P_m$ and a smoothness term or a similarity term $E_s(\tilde{V}_T)$ for vertices $V$ that aims to preserve local image structures. Please refer to ~\cite{Liu:2009} for the derivation of those two terms.
\noindent We then compute the final energy term as

\vspace{-0.10in}
\begin{equation}
    E=\alpha\sum_m{E_{p,m}}+\sum_T{E_s(\tilde{V}_T)}
\end{equation}
\vspace{-0.10in}

\noindent where $\alpha$ is the weight for the data term with a default value 0.5 for features in homogeneous regions and 1 for features at edge points. The minimization of $E$ is solved by constructing a linear system and solving it using standard sparse linear solver. The final warping result is rendered using texture mapping according to the output mesh.

\vspace{-0.05in}
\section{Labeling-based Frame Rendering}
\label{sec:method_blending}
\vspace{-0.05in}
The three input frames are now warped to the same temporal position. However, directly blend them together might introduce visible visual artifacts because warping holes and mis-matches still exist. For each rendering pixel in the final result, a subset selection of its three warped pixels (denoted as $p^s_k,p^r_{k-1},p^r_{k+1}$) needs to be made. As there are $2^3=8$ combinations of selections (as shown is Table~\ref{tab:renderinglabels}) for each rendering pixel, we formulate the decision making of all rendering pixels as a labeling problem, where each pixel is to be assigned one of the 8 labels, where each label $L=(l^s_k,l^r_{k-1},l^r_{k+1})$ contains three binary numbers indicating the selection of each warped pixel. Note that for pixels assigned with label 1 with $L=(0,0,0)$, leading to holes in the final rendering results, we use Poisson image inpainting~\cite{poissonImageEditing} with zero gradient to infill them.

\begin{table}[t]
\centering{
%\small
\footnotesize{
\caption{Eight labels for each rendering pixel}
\vspace{-0.10in}
\begin{tabular}{|c|c|c|c|c|}
    \hline
    label No. & 1     & 2        & 3        & 4    \\
    \hline
    notation  & $(0,0,0)$ & $(1,0,0)$ & $(0,1,0)$ & $(0,0,1)$\\
    \hline
    selection & none & $p^s_k$ & $p^r_{k-1}$ & $p^r_{k+1}$ \\
    \hline
    label No.  & 5     & 6        & 7        & 8   \\
    \hline
    notation  & $(0,1,1)$ & $(1,0,1)$ & $(1,1,0)$ & $(1,1,1)$\\
    \hline
    selection & $p^r_{k-1},p^r_{k+1}$ & $p^s_k,p^r_{k+1}$ & $p^s_k,p^r_{k-1}$ & all \\
    \hline
\end{tabular}
\label{tab:renderinglabels}
}
}
\vspace{-0.10in}
\end{table}
We consider the final rendered frame as an un-directed graph in which each rendering pixel is represented as a node and each pair of spatially neighbouring pixels are connected by an un-directed edge. This labeling problem can then be effectively solved using a graph cuts based multi-label energy minimization technique from Fulkerson~\etal~\cite{graphcuts}. Given the optimized labels for each pixel, we weighted average the corresponding selected subset of three warped pixels using optical flow validated weights $W$.

\vspace{0.05in}
\noindent \textbf{Labeling Data Term.} Following a statistic rule that more similar samples lead to better reconstruction results, we define the labeling data term for all rendering pixels ($p$) as

\vspace{-0.100in}
\begin{equation}
     E^L_{data} = \sum_p{(E^{L,p}_{ind}+\beta_L\cdot E^{L,p}_{sim})\cdot Z^{-1}_{L,p}}
\end{equation}
\vspace{-0.100in}

\noindent where $Z_{L,p}$ is a normalizing factor that encourages more selected samples and is defined as

\vspace{-0.100in}
\begin{equation}
     Z_{L,p} = (l^s_k+l^r_{k-1}+l^r_{k+1}+\epsilon)^{\alpha_L}_p
\end{equation}
\vspace{-0.100in}

\noindent where $\epsilon$ is a small constant ($\epsilon\ll1$). $E^{L,p}_{ind}$ gives credits to each individual selected pixels and is defined as

\vspace{-0.100in}
\begin{equation}
     E^{L,p}_{ind} = (K^s_{k} \cdot l^s_k + K^r_{k-1} \cdot l^r_{k-1} + K^r_{k+1} \cdot l^r_{k-1})_p
\end{equation}
\vspace{-0.100in}

\noindent where $K^s_{k}=1$ and $K^r_{k-1}=K^r_{k+1}=1.5$. They are parameters that control the weights for pixels from source frames and reference frames. We set a smaller $K^s_{k}$ value than the other two to prefer single selection from the source frames to reference frames. This is because scenes warped from the source frame is captured at the same time as the interpolating time stamp.

$E^{L,p}_{sim}$ is a similarity measuring term that penalizes large intensity difference between two pixels if both of them are selected. We thus define $E^{L,p}_{sim}$ as follows.

\vspace{-0.150in}
\begin{equation}
    \begin{aligned}
     E^{L,p}_{sim} = &l^s_k l^r_{k-1} d(p^s_k,p^r_{k-1}) + l^s_k\cdot l^r_{k+1} d(p^s_k,p^r_{k+1}) + \\  &l^r_{k-1}\cdot l^r_{k+1} d(p^r_{k-1},p^r_{k+1}) 
    \end{aligned}
\end{equation}
\vspace{-0.150in}

\noindent where $d(a,b)$ is the $l2$ norm difference of two pixels $a$ and $b$. This term encourages that only similar pixels are preferred to be added to the final selection. $\alpha_L$ and $\beta_L$ are two controlling parameters with default values of 3 and 8.

Note that for a pixel $p$, it could be possible that not all 8 labels are valid due to the existence of warping holes. For example, if no pixel is warped to some location in the warped source frame, then pixel $p^s_k$ can not be selected at this location as it does not exist. Thus, labels $(1,0,0), (1,0,1), (1,1,0)$ and $(1,1,1)$ are all invalid labels. For these invalid labels, we directly assign a large labeling data term to avoid invalid label selection $E^{L,p}_{d} = \infty$.

%\begin{equation}
%    E^{L,p}_{d} = \infty
%\end{equation}

\vspace{0.02in}
\noindent \textbf{Labeling Smoothness Term.} Neighboring pixels are more likely to have the same labels. We thus define the smoothness term as the $l2$ norm of label differences.

\vspace{-0.100in}
\begin{equation}
     E^{L,p,q}_{smooth} = ||L_p-L_q||^2
\end{equation}
\vspace{-0.100in}

\noindent where $L_p$ and $L_q$ are two labels for a pair of connected rendering pixels. The final smoothness term for all neighboring pixels are then defined as follows.

\vspace{-0.100in}
\begin{equation}
     E^L_{smooth} = \sum_{p,q}{E^{L,p,q}_{smooth}}
\end{equation}
\vspace{-0.100in}

The final energy function is then defined as 

\vspace{-0.100in}
\begin{equation}
     E^L = E^L_{data}+\gamma_L \cdot E^L_{smooth}
\end{equation}
\vspace{-0.100in}

\noindent where $\gamma_L$ is a controlling parameter with default value 2. After all labels are obtained via optimization, each of the final rendering pixels can be computed as a weighted average of all selected warped pixels.

\begin{figure}[t]
\vspace{-0.025in}
    \begin{center}
      \begin{tabular}{c}
		\hspace{-0.075in}\includegraphics[width=.42\textwidth]{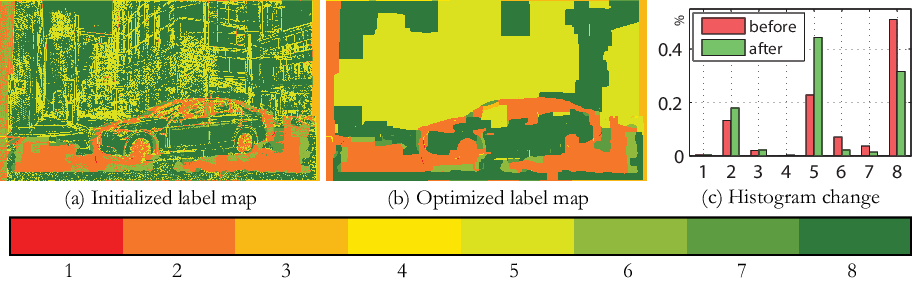}       
      \end{tabular}
      \vspace{-0.200in}
      \caption{Graph cuts-based labeling. (a): Initialized label map. (b): The final optimized label map by our method. (c): Label histogram comparison before and after optimization.\label{fig:labelingDemo}}
    \end{center}
	\vspace{-0.350in}
\end{figure}

We first initialize the label map for each pixel by selecting the label that minimizes the data term at current position, as shown in Figure~\ref{fig:labelingDemo} (a). We then use graph cuts multi-label optimization~\cite{graphcuts} to get the final label map. 

As the source and reference frames compliment each other, the optimization allows our approach to make good use of them. The warped source frames capture what is really going on at the current time stamp. They thus have better quality in occluded/dis-occluded or blurred regions. As shown in Figure~\ref{fig:labelingDemo} (b), in most occluded/dis-occluded regions, our labeling optimization selects pixels only from the source frames (indicating the selection of label 2 in Table~\ref{tab:renderinglabels}). However, as the source frames are imaged from a slightly different view point to the reference frames, they often suffer from parallax jittering effects. Thus, the warped reference frames have generally better qualities in such regions. From Figure~\ref{fig:labelingDemo}, it can be seen that in most regions in the background, the combination of pixels from the two reference frames are preferred (indicating the selection of label 5 in Table~\ref{tab:renderinglabels}). The fusion of the two types of frames thus makes our approach robust against various types of scenes. Statistically, it can be seen from Figure~\ref{fig:labelingDemo} (c) that our optimization replaces part of label 8 (selection of all three pixels) with label 5. Overall, our optimization does not change the labels' distribution significantly while preserving more neighbouring smoothness.

The warping of all input frames can create holes due to the existence of dis-occluded regions. We fill these holes using Poisson image inpainting~\cite{poissonImageEditing}. We follow the approach from Chaurasia~\etal~\cite{sigraph13warp} to assign zero gradients to these pixels for inpainting.

\vspace{-0.05in}
\section{Results}
\vspace{-0.05in}
\label{sec:exp}
\begin{figure}[tb]
\footnotesize{
    \begin{center}
      \begin{tabular}{ccccc}
        \hspace{-0.10in}\includegraphics[width=.092\textwidth]{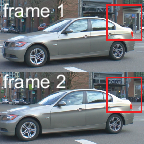}                       &
        \hspace{-0.15in}\includegraphics[width=.092\textwidth]{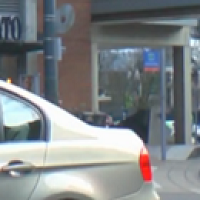}          &
        \hspace{-0.15in}\includegraphics[width=.092\textwidth]{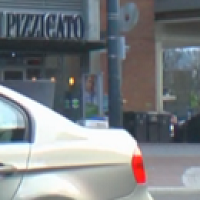}          &
		\hspace{-0.15in}\includegraphics[width=.092\textwidth]{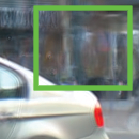}           &
		\hspace{-0.15in}\includegraphics[width=.092\textwidth]{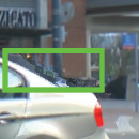}        \\ [-0.9ex]		
		\hspace{-0.10in}   silvercar   & \hspace{-0.15in}  Frame 1  & \hspace{-0.15in}  Frame 2  & \hspace{-0.15in} PHI & \hspace{-0.15in} SPF       \\ [-0.3ex]
		\hspace{-0.10in}\includegraphics[width=.092\textwidth]{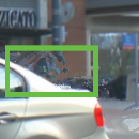}            &
        \hspace{-0.15in}\includegraphics[width=.092\textwidth]{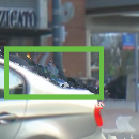}         &
		\hspace{-0.15in}\includegraphics[width=.092\textwidth]{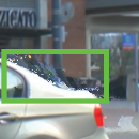}        &
		\hspace{-0.15in}\includegraphics[width=.092\textwidth]{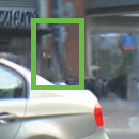}           &
		\hspace{-0.15in}\includegraphics[width=.092\textwidth]{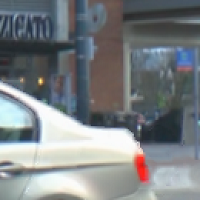}             \\ [-0.9ex]		
		\hspace{-0.10in} BRF  &  \hspace{-0.15in} CMP &  \hspace{-0.15in} DEEP &  \hspace{-0.15in} ASCNN &  \hspace{-0.15in} Ours                 \\ [0.2ex]
		
		\hspace{-0.10in}\includegraphics[width=.092\textwidth]{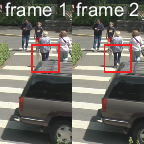}                       &
        \hspace{-0.15in}\includegraphics[width=.092\textwidth]{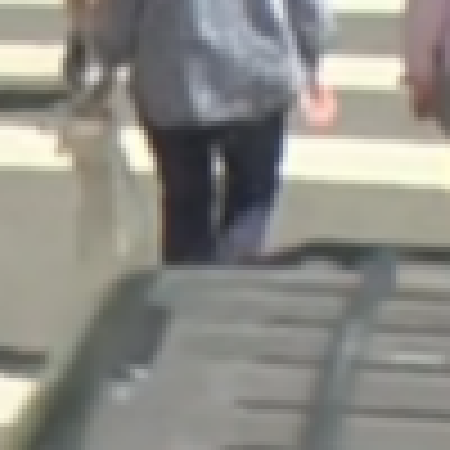}          &
        \hspace{-0.15in}\includegraphics[width=.092\textwidth]{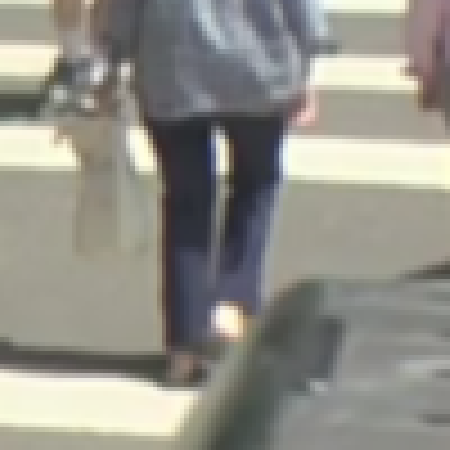}          &
		\hspace{-0.15in}\includegraphics[width=.092\textwidth]{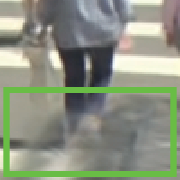}           &
		\hspace{-0.15in}\includegraphics[width=.092\textwidth]{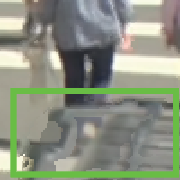}        \\[-0.9ex]		
		\hspace{-0.10in}   zebraped  & \hspace{-0.15in}  Frame 1  & \hspace{-0.15in}  Frame 2  & \hspace{-0.15in} PHI & \hspace{-0.15in} SPF        \\ [-0.3ex]
		\hspace{-0.10in}\includegraphics[width=.092\textwidth]{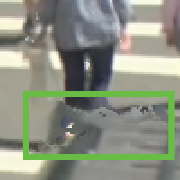}            &
        \hspace{-0.15in}\includegraphics[width=.092\textwidth]{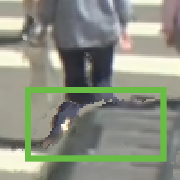}         &
		\hspace{-0.15in}\includegraphics[width=.092\textwidth]{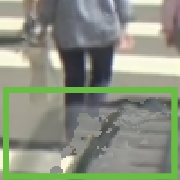}        &
		\hspace{-0.15in}\includegraphics[width=.092\textwidth]{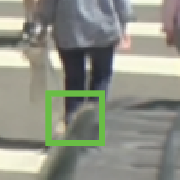}           &
		\hspace{-0.15in}\includegraphics[width=.092\textwidth]{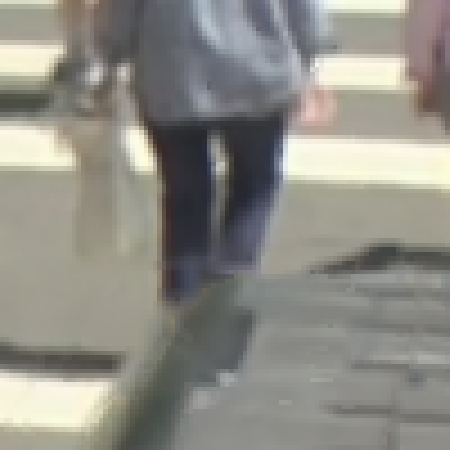}             \\ [-0.9ex]		
		\hspace{-0.10in} BRF  &  \hspace{-0.15in} CMP &  \hspace{-0.15in} DEEP &  \hspace{-0.15in} ASCNN &  \hspace{-0.15in} Ours                 \\ [0.2ex]
		
		\hspace{-0.10in}\includegraphics[width=.092\textwidth]{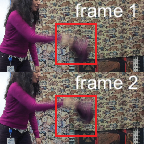}                       &
        \hspace{-0.15in}\includegraphics[width=.092\textwidth]{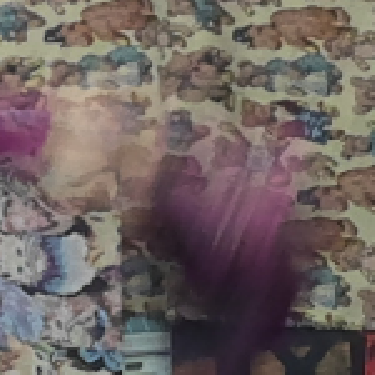}          &
        \hspace{-0.15in}\includegraphics[width=.092\textwidth]{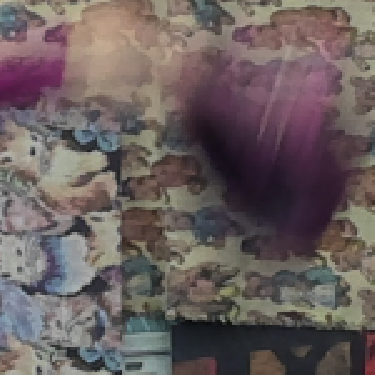}          &
		\hspace{-0.15in}\includegraphics[width=.092\textwidth]{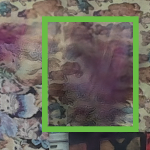}           &
		\hspace{-0.15in}\includegraphics[width=.092\textwidth]{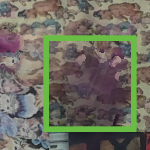}        \\[-0.9ex]		
		\hspace{-0.10in}   throwhat  & \hspace{-0.15in}  Frame 1  & \hspace{-0.15in}  Frame 2  & \hspace{-0.15in} PHI & \hspace{-0.15in} SPF        \\ [-0.3ex]
		\hspace{-0.10in}\includegraphics[width=.092\textwidth]{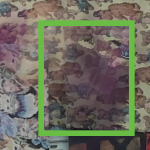}            &
        \hspace{-0.15in}\includegraphics[width=.092\textwidth]{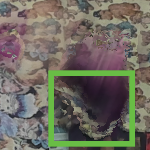}         &
		\hspace{-0.15in}\includegraphics[width=.092\textwidth]{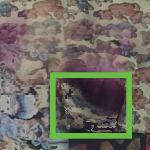}        &
		\hspace{-0.15in}\includegraphics[width=.092\textwidth]{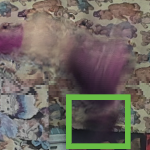}           &
		\hspace{-0.15in}\includegraphics[width=.092\textwidth]{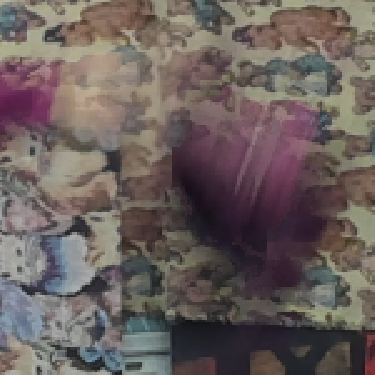}             \\ [-0.9ex]		
		\hspace{-0.10in} BRF  &  \hspace{-0.15in} CMP &  \hspace{-0.15in} DEEP &  \hspace{-0.15in} ASCNN &  \hspace{-0.15in} Ours                 \\ [0.2ex]
      \end{tabular}
	  \vspace{-0.16in}
      \caption{Comparison to single-lens interpolation methods. \label{fig:visual_02}}
    \end{center}
}
\vspace{-0.25in}
\end{figure}

We evaluate our approach using videos from RMIT3DV dataset~\cite{cheng2012rmit3dv}, Choubassi et al.~\cite{intel2017}, adtv.at  and videos captured by our own cameras as well as simulated videos generated using Maya 2016. Choubassi's dataset consisting of videos captured using 2 by 2 camera arrays, the Maya-simulated videos are captured using a virtual 3 by 3 camera array and all other videos are captured using 2 by 1 camera arrays. These videos contain a wide variety of scenes, including indoor and outdoor scenes with various levels of motion. There are also challenging scenes with parallax, large camera motion and blurred moving objects.

\begin{figure}[t]
    \vspace{-0.15in}
    \begin{center}
      \begin{tabular}{c}
		\hspace{-0.10in}\includegraphics[width=.40\textwidth]{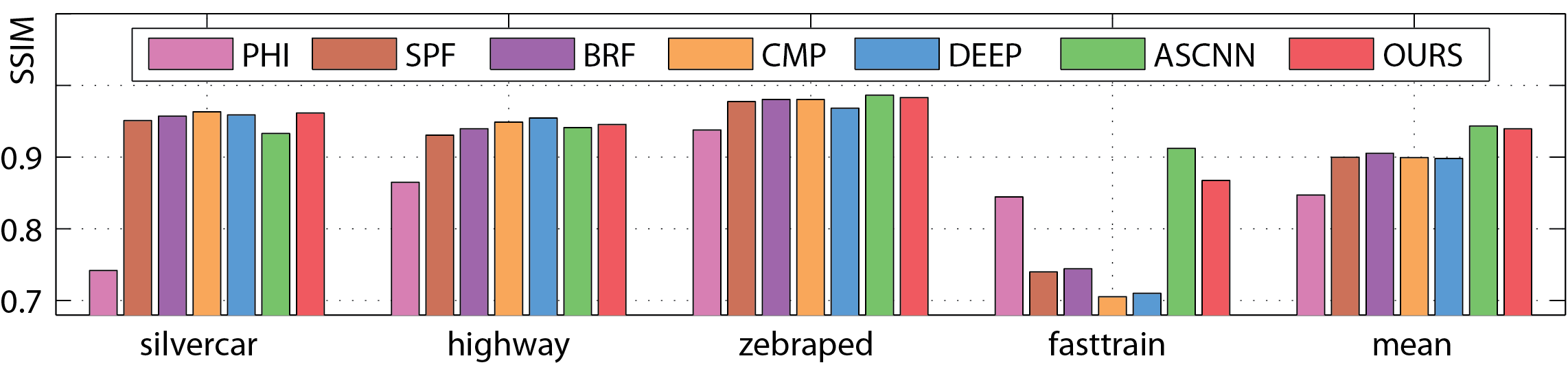}
      \end{tabular}
	  \vspace{-0.20in}
      \caption{Quantitative comparison to single-lens methods.\label{fig:quant_comp_chart}}
    \end{center}
	\vspace{-0.25in}
\end{figure}

\begin{figure}[b]
\vspace{-0.15in}
    \begin{center}
      \begin{tabular}{c}
		\hspace{-0.10in}\includegraphics[width=.40\textwidth]{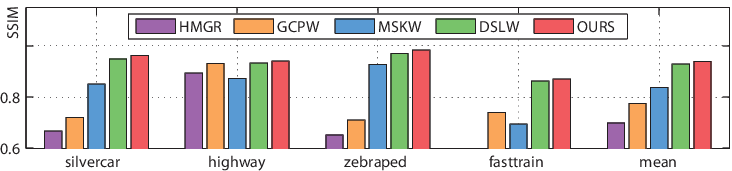}
      \end{tabular}
	  \vspace{-0.20in}
      \caption{Quantitative comparison to warp-based methods.\label{fig:quant_comp_chart_warp}}
    \end{center}
	\vspace{-0.15in}
\end{figure}

%\vspace{0.1in}
\noindent \textbf{Comparison with single-lens frame interpolation.} We compare our method to six recent single-lens frame interpolation methods. We compare to Phase-Based method (PHI)~\cite{meyer2015phase} and Adaptive Separable Convolution (ASCNN)~\cite{niklaus2017video} using the authors' implementations. We also compare our method to optical flow-based methods, including SparseFlow (SPF)~\cite{ijcv2011opf}, BroxFlow (BRF)~\cite{brox2004high}, CMPFlow (CMP)~\cite{hu2016efficient}  and DeepFlow (DEEP)~\cite{weinzaepfel2013deepflow}. For all optical flow algorithms, we use the codes provided by the authors. To interpolate the in-between images given the estimated flow fields, we use the code provided by the author of the Middlebury interpolation benchmark~\cite{flow_interp_09}. 

\begin{figure}[tb]
\vspace{-0.10in}
\footnotesize{
    \begin{center}
      \begin{tabular}{cccc}
      
        \hspace{-0.08in}\includegraphics[width=.100\textwidth]{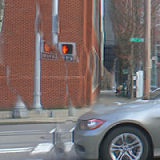}          &
        \hspace{-0.15in}\includegraphics[width=.100\textwidth]{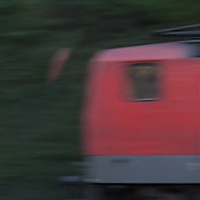}          &
		\hspace{-0.15in}\includegraphics[width=.100\textwidth]{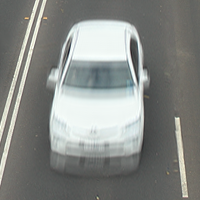}            &
		\hspace{-0.15in}\includegraphics[width=.100\textwidth]{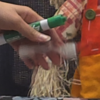}            \\
		\hspace{-0.08in}\includegraphics[width=.100\textwidth]{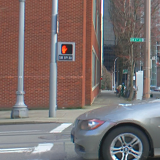}            &
        \hspace{-0.15in}\includegraphics[width=.100\textwidth]{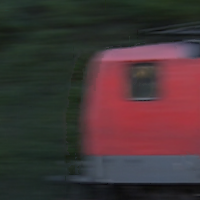}            &
		\hspace{-0.15in}\includegraphics[width=.100\textwidth]{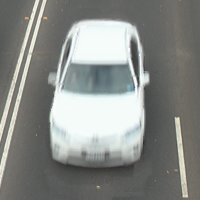}            &
		\hspace{-0.15in}\includegraphics[width=.100\textwidth]{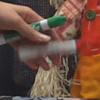}            \\  
      \end{tabular}
      \vspace{-0.15in}
      \caption{Visual comparison between the ASCNN~\cite{niklaus2017video} (top row) and our approach (bottom row) on challenging scenes with large motion and blurry moving objects. \label{fig:visual_comptoCNN}}
    \end{center}
}
\vspace{-0.25in}
\end{figure}

In Figure~\ref{fig:visual_02} we visually compare our interpolation results on scenes with large motion or blurred moving objects. PHI introduces additional blur to moving contents as high frequency contents cannot be represented by phase estimation. Optical flow-based methods produce distortions at occluded/dis-occluded regions. Optical flow-based methods also fail to interpolate blurred moving objects. They tend to blend the foreground and background as they ignore the blurred moving objects and consider them as static. CMP and DEEP produce better optical flow estimations. However, they introduce serious artifacts at moving boundaries. In contrast, our approach can generate visual plausible results by invalidating the guidance of incorrect flow estimations in the local warp and let nearby superpixels to help with the warp. ASCNN can generate good results for scenes with small or moderate motion and occlusion. However, for scenes with large motion (as shown in \textit{slivercar} in Figure~\ref{fig:visual_02}) or occlusion (as shown in \textit{throwhat} in Figure~\ref{fig:visual_02}), ASCNN introduces noticeable visual artifacts while our approach can still generate plausible interpolated frames for those challenging scenes. This observation can also be confirmed by results shown in Figure~\ref{fig:visual_comptoCNN}.

\begin{figure}[tb]
\vspace{-0.05in}
\footnotesize{
    \begin{center}
      \begin{tabular}{cccc}
      
        \hspace{-0.08in}\includegraphics[width=.100\textwidth]{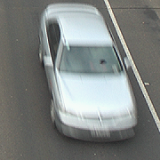}          &
        \hspace{-0.15in}\includegraphics[width=.100\textwidth]{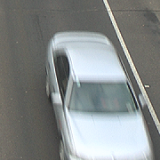}          &
		\hspace{-0.15in}\includegraphics[width=.100\textwidth]{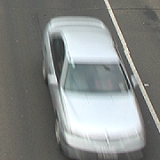}            &
		\hspace{-0.15in}\includegraphics[width=.100\textwidth]{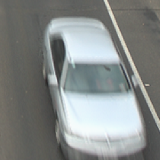}            \\ [-0.7ex]		
		\hspace{-0.08in} Input 1  & \hspace{-0.15in}  Input 2  & \hspace{-0.15in} HMGR & \hspace{-0.15in} GCPW               \\ %[-0.3ex]
		\hspace{-0.08in}\includegraphics[width=.100\textwidth]{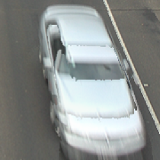}            &
        \hspace{-0.15in}\includegraphics[width=.100\textwidth]{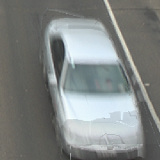}            &
		\hspace{-0.15in}\includegraphics[width=.100\textwidth]{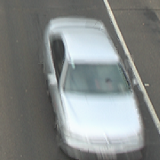}            &
		\hspace{-0.15in}\includegraphics[width=.100\textwidth]{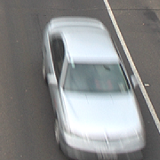}              \\ [-0.7ex]		
		\hspace{-0.08in} MSKW  &  \hspace{-0.15in} DSLW &  \hspace{-0.15in} OURS &  \hspace{-0.15in} GT                      \\ %[-0.3ex]
		
		\hspace{-0.08in}\includegraphics[width=.100\textwidth]{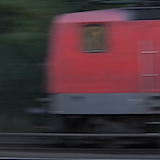}          &
        \hspace{-0.15in}\includegraphics[width=.100\textwidth]{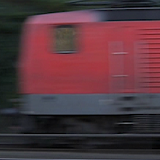}          &
		\hspace{-0.15in}\includegraphics[width=.100\textwidth]{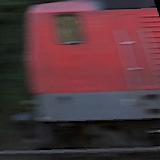}            &
		\hspace{-0.15in}\includegraphics[width=.100\textwidth]{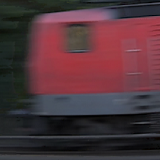}            \\ [-0.7ex]	
		\hspace{-0.08in} Input 1  & \hspace{-0.15in}  Input 2  & \hspace{-0.15in} HMGR & \hspace{-0.15in} GCPW               \\ %[-0.3ex]
		\hspace{-0.08in}\includegraphics[width=.100\textwidth]{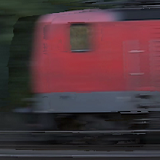}            &
        \hspace{-0.15in}\includegraphics[width=.100\textwidth]{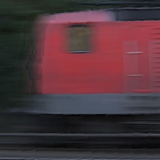}            &
		\hspace{-0.15in}\includegraphics[width=.100\textwidth]{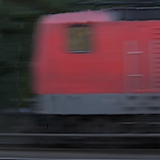}            &
		\hspace{-0.15in}\includegraphics[width=.100\textwidth]{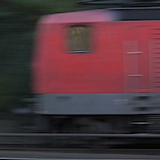}              \\ [-0.7ex]	
		\hspace{-0.08in} MSKW  &  \hspace{-0.15in} DSLW &  \hspace{-0.15in} OURS &  \hspace{-0.15in} GT                      \\ %[-0.3ex]
      \end{tabular}
	  \vspace{-0.20in}
      \caption{Visual comparison to warp-based methods. \label{fig:visual_warp}}
    \end{center}
}
\vspace{-0.25in}
\end{figure}

We quantitatively test our method on 4 videos (as shown in Figure~\ref{fig:visual_02} and Figure~\ref{fig:visual_warp}) with ground truth, which are obtained using the leave-some-out method. Specifically, we interpolate intermediate frames and compare them to the original ones. We report the perceptually motivated structural similarity (SSIM) in Figure~\ref{fig:quant_comp_chart}. We also report the Mean Square Error (MSE) in the supplementary video due to space limit. In general, our approach has comparable quantitative performance to ASCNN and outperforms other competing methods. As shown in Figure~\ref{fig:visual_02} and Figure~\ref{fig:visual_comptoCNN}, Our method performs significantly better in challenging places such as occluded regions, blurry and fast moving objects and regions with large parallax. While those regions often occupy a small portion of the scenes, leading to limited overall quantitative improvements, they have large impact on the visual qualities, please refer to our supplementary material for details.

\begin{figure}[t]
\vspace{-0.05in}
\footnotesize{
    \begin{center}
      \begin{tabular}{c}
	     \includegraphics[width=.45\textwidth]{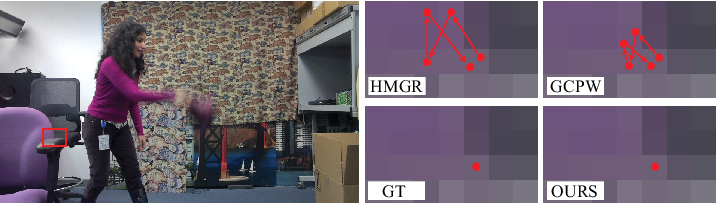}
      \end{tabular}
	  \vspace{-0.175in}
      \caption{HMGR fails to align the static background features. GCPW performs better, but still suffers from moderate parallax jittering. Our result properly maintains temporal coherence according to the ground truth (GT).\label{fig:warp_comp_traj}}
    \end{center}
}
\vspace{-0.20in}
\end{figure}

\noindent \textbf{Comparison with warp-based methods.} We compare our method to four warp-based methods, include homography transformation (HMGR), global content-preserving warp (GCPW)~\cite{Liu:2009}, depth synthesis and local warps (DSLW)~\cite{sigraph13warp} and mask-based warps (MSKW) from Choubassi et al.~\cite{intel2017}. We show the visual comparison in Figure~\ref{fig:visual_warp}. Comparing to DSLW and MSKW, our approach generates interpolating results with less visual artifacts (duplication/blur) as our approach effectively eliminates bad optical flow for warping guidence. While HMGR and GCPW tend to have less visual artifacts as they globally warp the source frames to the reference, they often suffer from parallax jittering, as can be seen in the second example in Figure~\ref{fig:visual_warp}. To further verify this we compare our method to those two approaches by tracking static feature trajectories across consecutive interpolated frames in Figure~\ref{fig:warp_comp_traj}. It can be seen that the trajectory in our result better preserves temporal coherence and is closer to the ground truth trajectory (GT). The difference becomes more obvious in the supplementary video. The plot in Figure~\ref{fig:quant_comp_chart_warp} shows that our approach quantitatively performs better than other warp-based interpolation methods. This observation can also be confirmed by the MSE comparison in our supplementary video.

\begin{figure}[t]
%\vspace{-0.00in}
\footnotesize{
    \begin{center}
      \begin{tabular}{c}
		\hspace{-0.10in}\includegraphics[width=.40\textwidth]{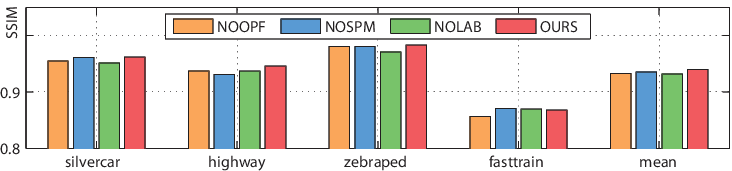}
      \end{tabular}
	  \vspace{-0.175in}
      \caption{Quantitative leave-one-out component analysis.\label{fig:leave_comp_out_stat}}
    \end{center}
}
\vspace{-0.25in}
\end{figure}

\begin{figure}[t]
\vspace{-0.05in}
\footnotesize{
    \begin{center}
      \begin{tabular}{cccc}
		\hspace{-0.12in} \includegraphics[width=.100\textwidth]{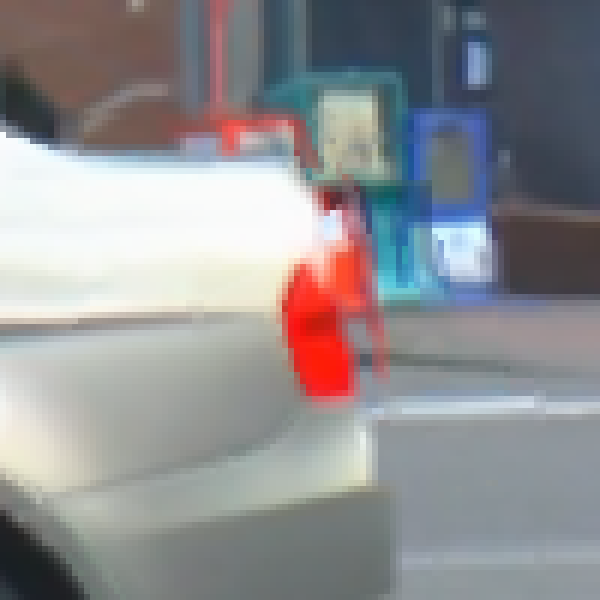}     &
        \hspace{-0.18in} \includegraphics[width=.100\textwidth]{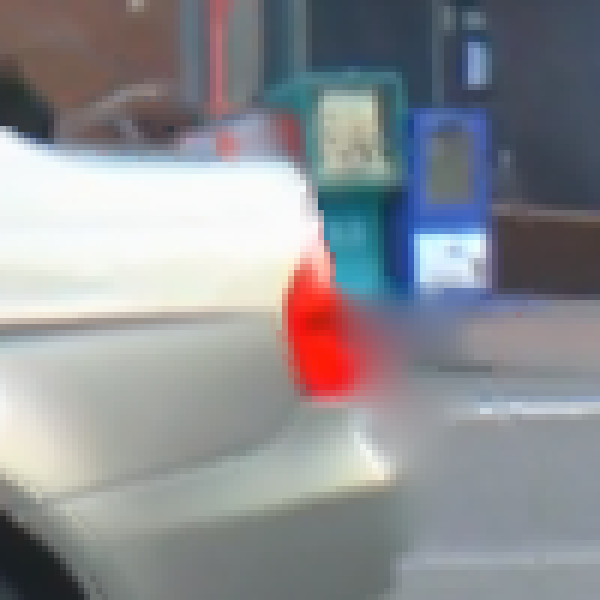}     &
        \hspace{-0.18in} \includegraphics[width=.100\textwidth]{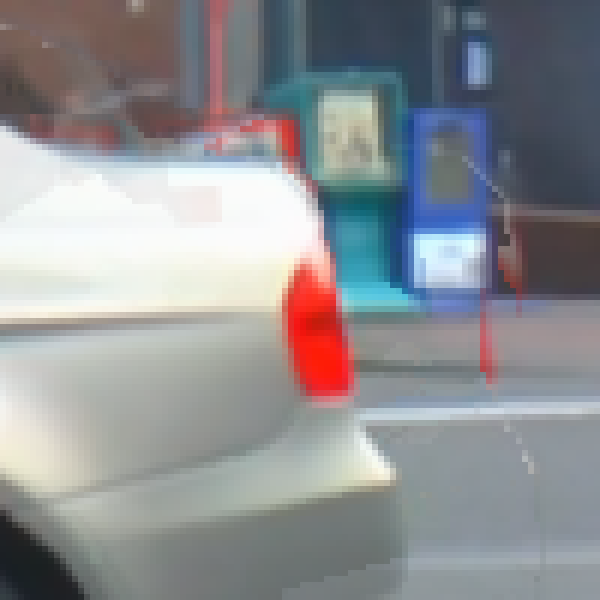}     &
        \hspace{-0.18in} \includegraphics[width=.100\textwidth]{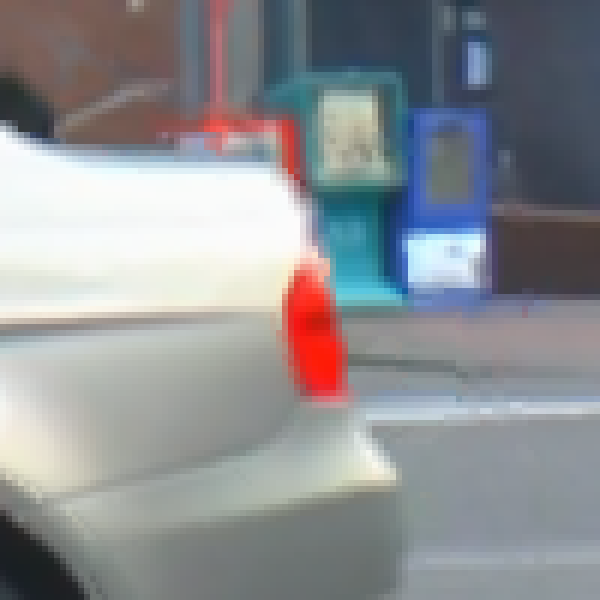}      \\ [-0.6ex]
        \hspace{-0.12in} (a) OPF$_{out}$  & \hspace{-0.15in} (b) SPM$_{out}$ & \hspace{-0.15in} (c) LAB$_{out}$  &  \hspace{-0.15in} (d) Our result
      \end{tabular}
	  \vspace{-0.15in}
      \caption{Leave-one-out evaluation of our method.\label{fig:leaveOneOut}}
    \end{center}
	\vspace{-0.30in}
}
\end{figure}

\noindent \textbf{Component analysis.} As our main contribution is an optical flow validation and a superpixel merging for local content preserving warp as well as a labeling-based frame rendering to blend multiple warped frames, we analyze these components by conducting leave-one-out experiments on them. Specifically, to leave the optical flow validation out (OPF$_{out}$), we set $W=1$ for all pixels. To leave out the superpixel merging (SPM$_{out}$), we simply skip it. To leave out labeling-based frame rendering, we replace it with a simple averaging scheme as used by Chaurasia~\etal ~\cite{sigraph13warp} (LAB$_{out}$). The plot in Figure~\ref{fig:leave_comp_out_stat} shows quantitative degradation when leaving out any of those components. For local content preserving warp, our optical flow validation effectively removes flow outliers and superpixel merging assigns reasonable good flow for bad superpixels from neighbouring regions. The following labeling-based frame rendering then makes the best use of all individually warped frames to attenuates errors by letting them complement each other. As shown in Figure~\ref{fig:leaveOneOut}, leaving out any component would introduce noticeable visual artifacts.

\noindent \textbf{Implementation} The proposed approach is implemented using C++ and MATLAB on a desktop with a 4-core Intel i7-4770 3.40GHz
CPU. This unoptimized off-line implementation takes an average of 87.34 seconds to synthesize a frame of size 720$\times$396.

\begin{figure}[tb]
\vspace{-0.06in}
\footnotesize{
    \begin{center}
      \begin{tabular}{ccc}
		\hspace{-0.08in}\includegraphics[width=.140\textwidth]{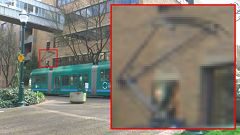}     &
        \hspace{-0.15in}\includegraphics[width=.140\textwidth]{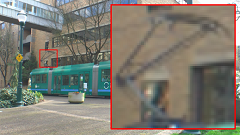}     &
        \hspace{-0.15in}\includegraphics[width=.140\textwidth]{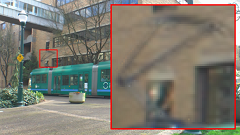}     \\
        
        \hspace{-0.11in} (a) Frame 1  & \hspace{-0.15in} (b) Frame 2 & \hspace{-0.15in} (c) Interpolated                  
      \end{tabular}
	  \vspace{-0.15in}
      \caption{Thin structures that can not full-fill a single superpixel is blurred by our method.\label{fig:thin_struc}}
    \end{center}
}
\vspace{-0.25in}
\end{figure}

\begin{figure}[tb]
\vspace{-0.02in}
\footnotesize{
    \begin{center}
      \begin{tabular}{ccc}

        \hspace{-0.08in}\includegraphics[width=.140\textwidth]{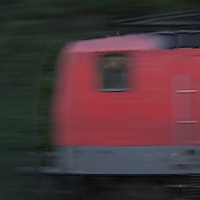}     &
        \hspace{-0.15in}\includegraphics[width=.140\textwidth]{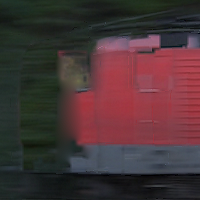}     &
        \hspace{-0.15in}\includegraphics[width=.140\textwidth]{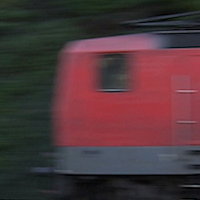}     \\
        
        \hspace{-0.11in} (a) 1 frame out  & \hspace{-0.15in} (b) 3 frames out & \hspace{-0.15in} (c) Ground truth                  
      \end{tabular}
	  \vspace{-0.15in}
      \caption{Our method can handle relatively large motion (a), but can still fail when the motion becomes too large(b).\label{fig:large_motion}}
    \end{center}
}
\vspace{-0.30in}
\end{figure}

\noindent \textbf{Discussion and limitations.} 
While our approach can generate plausible interpolated videos, it has some limitations. Although our approach can handle parallax for moderately small objects in different depth, it introduces some blurring when the foreground objects are too small to be covered by a single superpixel. As can be seen in Figure~\ref{fig:thin_struc}, the final interpolated antenna is blurred as the local warp is mainly guided by flow in the background regions in its corresponding superpixels. In addition, while our approach can deal with large motion, it can fail when the motion becomes too large. As shown in Figure~\ref{fig:large_motion} (a), our method is able to generate interpolated frame with reasonably good quality for scenes with a foreground motion of about -35 pixels and background motion of 70 pixels. However, when we double the motion by leaving more frames out to synthesize the input frames, noticeable visual artifacts occur in the interpolated result, as shown in Figure~\ref{fig:large_motion} (b).

\vspace{-0.05in}
\section{Conclusion}
\vspace{-0.05in}
\label{sec:con}

In this paper, we propose a warping-based method to generate high frame rate videos using an asynchronous low frame rate camera array in which the video frames are alternatively captured by each camera. We first over-segment the input frames into superpixels, we locally warp each individual superpixel from the source frames to the reference with the help of validated optical flow fields and modified superpixel maps in which superpixels with poor flow estimations are merged to nearby neighbours. By utilizing the fusion of both the current source frame and temporally neighbouring reference frames using a graph cuts-based optimization labeling, our approach can produce plausible high-speed videos with high qualities on a variety of scenes with different levels of motions.

\vspace{-0.05in}
\vspace{2mm} \hspace{-6mm} \textbf{Acknowledgements.} This project was in part supported by a gift award from Intel.
\vspace{-0.05in}

{\small
\bibliographystyle{ieee}
\bibliography{ref}
}

\end{document}